\documentclass[11pt]{article}

\usepackage[margin=1in]{geometry}
\usepackage{amsmath,amssymb,amsthm}
\usepackage{graphicx}
\usepackage{hyperref}
\usepackage{cleveref}
\usepackage{booktabs}
\usepackage{enumitem}
\usepackage{xcolor}
\usepackage{subcaption}
\usepackage{listings}
\usepackage{pifont}
\usepackage{natbib}
\hypersetup{
  colorlinks=true,
  linkcolor=blue!70!black,
  citecolor=green!50!black,
  urlcolor=blue!70!black
}

\newtheorem{theorem}{Theorem}

\lstdefinelanguage{Lean}{
  morekeywords={theorem,def,where,by,fun,let,have,show,sorry,import,open,namespace,end,if,then,else,do,return,match,with,inductive,structure,class,instance,example,noncomputable,variable,section,set_option},
  sensitive=true,
  morecomment=[l]{--},
  morecomment=[s]{/-}{-/},
  morestring=[b]",
  literate={ℝ}{{$\mathbb{R}$}}1 {ℕ}{{$\mathbb{N}$}}1 {→}{{$\to$}}1 {∀}{{$\forall$}}1 {∃}{{$\exists$}}1 {∧}{{$\wedge$}}1 {≤}{{$\leq$}}1 {‖}{{$\|$}}1 {∫}{{$\int$}}1 {•}{{$\bullet$}}1 {ν}{{$\nu$}}1 {ρ}{{$\rho$}}1 {₀}{{$_0$}}1,
}
\newcommand{\lean}{\textsc{Lean}}
\newcommand{\mathlib}{\textsc{Mathlib}}
\newcommand{\sorry}{\texttt{sorry}}
\newcommand{\babysit}{\texttt{/babysit}}
\newcommand{\RR}{\mathbb{R}}
\newcommand{\TT}{\mathbb{T}}

\title{Semi-Autonomous Formalization of the\\Vlasov-Maxwell-Landau Equilibrium}
\author{Vasily Ilin\thanks{University of Washington. Email: \texttt{vilin@uw.edu}. Repository: \url{https://github.com/Vilin97/aristotle/tree/landau}.}}

\date{March 2026}

\begin{document}
\maketitle

\begin{abstract}
We present a complete \lean~4 formalization of the equilibrium characterization in the Vlasov-Maxwell-Landau (VML) system, which describes the motion of charged plasma. The project demonstrates the full AI-assisted mathematical research loop: an AI reasoning model (Gemini DeepThink) generated the proof from a conjecture, an agentic coding tool (Claude Code) translated it into \lean\ from natural-language prompts, a specialized prover (Aristotle) closed 111 lemmas, and the \lean\ kernel verified the result. A single mathematician supervised the process over 10 days at a cost of \$200, writing zero lines of code.

The entire development process is public: all 229 human prompts, and 213 git commits are archived in the repository. We report detailed lessons on AI failure modes -- hypothesis creep, definition-alignment bugs, agent avoidance behaviors -- and on what worked: the abstract/concrete proof split, adversarial self-review, and the critical role of human review of key definitions and theorem statements. Notably, the formalization was completed before the final draft of the corresponding math paper was finished.
\end{abstract}

\begin{table}[t]
\centering
\footnotesize
\setlength{\tabcolsep}{3pt}
\begin{tabular}{@{}llrccccccl@{}}
\toprule
Project & Prover & LOC & New thm & New defs & Team & AI role & Logs & Domain \\
\midrule
Gauss~\citep{mathinc2026sphere} & \lean~4 & 181K\textsuperscript{$\ddagger$} & {\color{red!70!black}\ding{55}} & {\color{red!70!black}\ding{55}} & 6+\textsuperscript{$\S$} & sorry elim. & {\color{red!70!black}\ding{55}} & Geometry \\
Bayer \& David~\citep{davidbayer2025} & Isabelle & 20K & {\color{green!60!black}\ding{51}} & {\color{green!60!black}\ding{51}} & 15 & none & {\color{red!70!black}\ding{55}} & Num.\ theory \\
\textbf{This work} & \lean~4 & \textbf{10K} & {\color{green!60!black}\ding{51}} & {\color{green!60!black}\ding{51}} & \textbf{1} & \textbf{all code} & {\color{green!60!black}\ding{51}} & \textbf{Math.\ phys.} \\
Numina~\citep{numina2025} & \lean~4 & $\sim$8K & {\color{green!60!black}\ding{51}} & {\color{green!60!black}\ding{51}} & 13 & collab. & {\color{red!70!black}\ding{55}} & Analysis \\
AxiomProver~\citep{axiom2026partial} & \lean~4 & 4.3K & {\color{green!60!black}\ding{51}} & $\sim$ & 21 & fully auton. & {\color{red!70!black}\ding{55}} & Num.\ theory \\
AxiomProver~\citep{axiom2026fel} & \lean~4 & 2.7K & {\color{green!60!black}\ding{51}} & $\sim$ & 21 & fully auton. & {\color{red!70!black}\ding{55}} & Algebra \\
2HDM~\citep{toobysmith2026} & \lean~4 & 1.6K & {\color{red!70!black}\ding{55}}\textsuperscript{$\dagger$} & {\color{green!60!black}\ding{51}} & 1 & minimal & {\color{red!70!black}\ding{55}} & Physics \\
Aristotle~\citep{erdos728} & \lean~4 & 1.4K & {\color{green!60!black}\ding{51}} & {\color{red!70!black}\ding{55}} & 1 & fully auton. & {\color{red!70!black}\ding{55}} & Combin. \\
AxiomProver~\citep{axiom2026parity} & \lean~4 & 0.2K & {\color{green!60!black}\ding{51}} & $\sim$ & 5 & fully auton. & {\color{red!70!black}\ding{55}} & Alg.\ geom. \\
\bottomrule
\end{tabular}
\caption{Recent research-level formalizations with publicly available code, sorted by LOC. ``New thm'' = theorem previously unpublished. ``New defs'' = required definitions or theory not in the proof library ({\color{green!60!black}\ding{51}}~= substantial, $\sim$~= lightweight, {\color{red!70!black}\ding{55}}~= none). ``AI role'': sorry elim.\ = eliminated sorry's in human-written scaffold; all code = AI generated all proof-assistant code from NL prompts; collab.\ = human-AI pair programming; fully auton.\ = AI from NL problem statement alone. ``Logs'' = full development process public (all commits, prompts, and interaction logs). \textsuperscript{$\dagger$}Discovered an error in the published proof. \textsuperscript{$\ddagger$}$\sim$22K lines (definitions, statements) written by human experts; Gauss eliminated $\sim$160K lines of sorry's in 3 weeks. \textsuperscript{$\S$}No paper; 6+ named contributors on blog post.}
\label{tab:comparison}
\end{table}

\section{Introduction}

Formal verification of mathematics in proof assistants like \lean~4 \citep{lean4,mathlib2020} has traditionally required deep expertise in both the mathematics and the proof assistant itself. The learning curve is steep: even experienced mathematicians report months of effort to formalize results that take pages to prove on paper \citep{urban2025topology}. This bottleneck has limited formal verification to a small community of specialists.

\begin{figure}[t]
  \centering
  \includegraphics[width=0.85\textwidth]{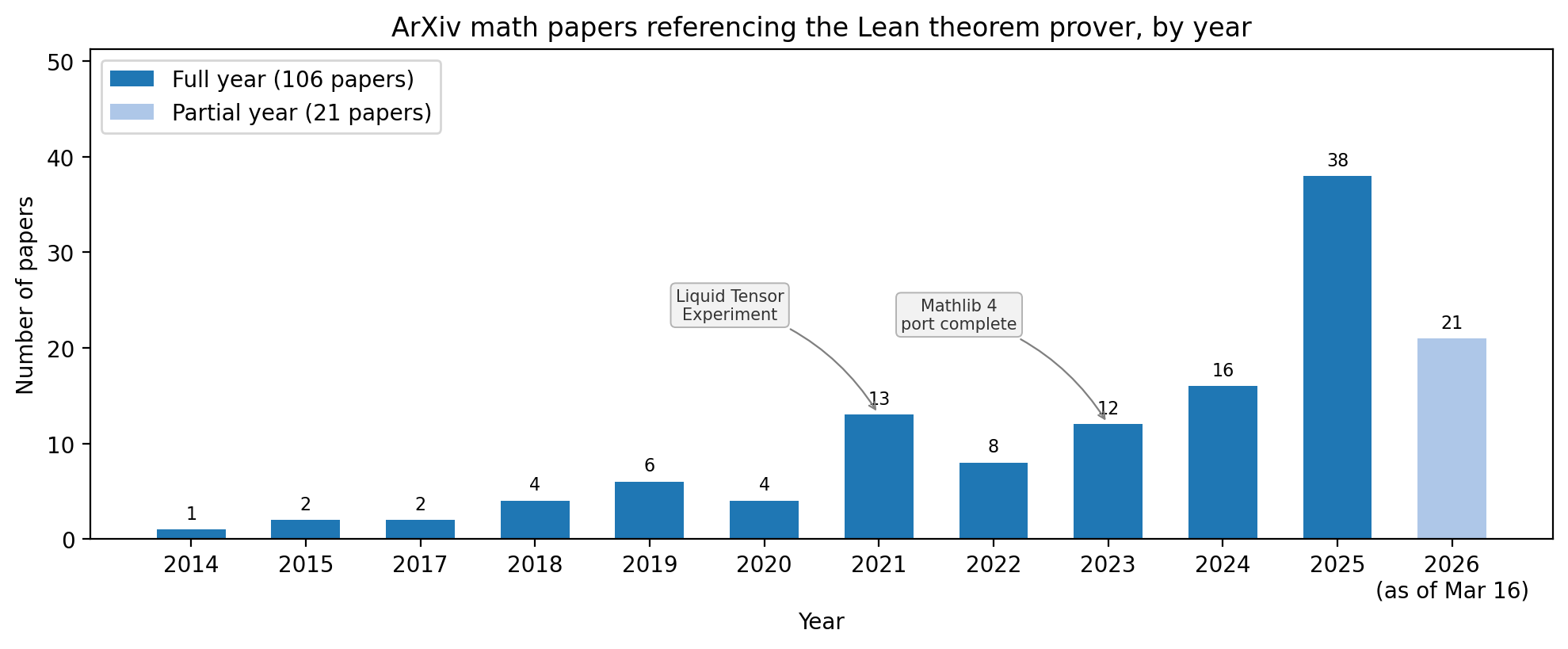}
  \caption{ArXiv papers using \lean\ for formalization, by year. Data from the arXiv metadata snapshot (March 2026), filtered for references to the \lean\ theorem prover.}
  \label{fig:arxiv-lean}
\end{figure}

Recent advances in AI-assisted theorem proving are changing this landscape rapidly (\Cref{fig:arxiv-lean}). Large language models can now generate \lean\ tactics \citep{seedprover2025}, agentic coding tools can manage entire \lean\ projects \citep{numina2025}, and cloud-based automated theorem provers can close individual lemmas without human intervention \citep{aristotle2025}.

Yet most published work on AI-assisted formalization focuses on competition mathematics (IMO, Putnam) or textbook results. Recent breakthroughs include AxiomProver autonomously generating a formal \lean\ proof of Fel's open conjecture \citep{axiom2026fel} from a natural-language problem statement alone, GPT-5.2 generating informal proofs of open Erd\H{o}s problems that were then auto-formalized in \lean\ by Aristotle \citep{erdos728}, and Google DeepMind's Aletheia agent \citep{aletheia2026,erdos_gemini2026} autonomously producing research papers in natural language. \citet{numina2025} formalized effective Brascamp-Lieb inequalities ($\sim$8K lines) with human experts collaborating on the \lean\ code, and \citet{davidbayer2025} formalized their own new theorem in Isabelle in parallel with the mathematical research, writing the proof assistant code themselves.

Our work is distinguished not by full autonomy -- AxiomProver's Fel's conjecture result is more autonomous -- but by the combination of several features that, to our knowledge, no single prior project shares (\Cref{tab:comparison}):
\begin{itemize}[leftmargin=2em,topsep=4pt,itemsep=2pt]
  \item \textbf{Scale and depth.} 34 files and 10K+ lines of \lean, in hard analysis: Coulomb singularity cancellation, Leibniz integral rule, energy methods on the torus.
  \item \textbf{New theory.} The theorem cannot even be \emph{stated} with existing \mathlib\ -- the Landau operator, torus differential structure, and VML system had to be defined from scratch.
  \item \textbf{New result.} The characterization of smooth steady states of the full VML system with Coulomb collisions and electromagnetic coupling does not appear in the prior literature in this precise form and is, to our knowledge, new as a stated and proved result.
  \item \textbf{Fully documented process.} A single mathematician wrote zero code while actively steering the formalization over 10 days; every prompt, commit, and tool call is public.
\end{itemize}

The contributions are:
\begin{enumerate}[leftmargin=2em]
  \item A demonstration of the full AI-assisted mathematical research loop: from an open conjecture, through AI-generated resolution and detailed proof (Gemini DeepThink, requiring no mathematical corrections), to a complete machine-verified formalization (Claude Code + Aristotle) -- all with the human acting solely as supervisor. To our knowledge, this is the first project to close this entire pipeline on a research-level result in hard analysis.
  \item A complete \lean~4 formalization of a new theorem in mathematical physics (the VML steady-state characterization), verified with zero \sorry's.
  \item A detailed, fully transparent account of the process: every prompt, every commit, every tool call is public.
  \item Lessons learned about the capabilities and failure modes of current AI tools for formalization, including a definition-alignment bug caught by an expert reviewer post-completion.
  \item Evidence that research-level formalization is now accessible to mathematicians without deep \lean\ expertise, at minimal cost (\$200 subscription + \$0 for Aristotle and Gemini).
\end{enumerate}

\section{The Theorem}
\label{sec:theorem}

\subsection{The Physical System}

The Vlasov-Maxwell-Landau (VML) system is a system of PDEs modeling the dynamics of a charged plasma. The full time-dependent system evolves a particle distribution function $f(t,x,v)$, an electric field $E(t,x)$, and a magnetic field $B(t,x)$ via the Vlasov equation coupled to Maxwell's equations with the Landau collision operator.

We are interested in \emph{steady states}: solutions with no time dependence. Setting all time derivatives to zero, the unknowns reduce to $f(x,v) > 0$ on $\TT^3 \times \RR^3$, $E(x)$, and $B(x)$, and the system becomes:
\begin{align}
  v \cdot \nabla_x f + (E + v \times B) \cdot \nabla_v f &= \nu \, Q(f,f), \label{eq:vlasov} \\
  \nabla \times B &= J = \int_{\RR^3} v \, f \, dv, \label{eq:ampere} \\
  \nabla \cdot E &= \int_{\RR^3} f \, dv - \rho_{\mathrm{ion}}, \label{eq:gauss} \\
  \nabla \cdot B &= 0, \label{eq:divB}
\end{align}
where $Q(f,f)$ is the Landau collision operator with the Coulomb kernel $\Psi(r) = r^{-3}$. Note that Faraday's law $\partial_t B = -\nabla \times E$ becomes $\nabla \times E = 0$ at equilibrium; this is not listed as a hypothesis because it is derived in the proof (it follows from the other equations).

\subsection{Main Result}

\begin{theorem}[Informal; see {\citet{ilin2026vml}} for the complete proof]
\label{thm:main-informal}
Let $f > 0$ be a smooth steady-state solution of the VML system with Coulomb collisions on $\TT^3 = (\RR/\mathbb{Z})^3$, with Schwartz-class velocity decay and a polynomial log-growth bound. Then:
\begin{enumerate}
  \item $f$ is a spatially uniform Maxwellian: $f(x,v) = \frac{\rho_{\mathrm{ion}}}{(2\pi T)^{3/2}} \exp\!\bigl(-\tfrac{|v|^2}{2T}\bigr)$ for some $T > 0$;
  \item the electric field vanishes: $E(x) = 0$;
  \item the magnetic field is constant: $B(x) = B_0$.
\end{enumerate}
\end{theorem}

The formal \lean\ statement is shown in \Cref{fig:lean-statement}. It takes 12 hypotheses, all with clear physical meaning: 2 physical parameters ($\nu > 0$, $\rho_{\mathrm{ion}} > 0$), strict positivity of $f$, smoothness conditions ($C^3$ in velocity, $C^2$ in space via periodic lift, $C^2$ for $B$), Schwartz-class velocity decay, a polynomial score bound ($|\partial_{v_i} f| \leq C(1+\|v\|)^K f$), and the four steady-state equations~(1) -- (4). The polynomial score bound is a genuine assumption not derivable from Schwartz decay alone; it is satisfied by Maxwellians and physically reasonable perturbations. A 13th hypothesis -- a polynomial log-growth bound ($|\log f(x,v)| \leq C(1+\|v\|)^K$) -- is derived from the score bound and smoothness inside the proof.

\begin{figure}[t]
\begin{lstlisting}
theorem CoulombConcreteTheorem42
    (f : Torus3 → (Fin 3 → ℝ) → ℝ)
    (E B : Torus3 → Fin 3 → ℝ)
    (ν ρ_ion : ℝ)
    (hν : 0 < ν)                                             -- (1)
    (hρ_ion : 0 < ρ_ion)                                     -- (2)
    (hf_pos : ∀ x v, 0 < f x v)                             -- (3)
    (hf_smooth_v : ∀ x, ContDiff ℝ 3 (f x))                  -- (4)
    (hf_smooth_x : ∀ v, ContDiff ℝ 2
      (periodicLift (fun x => f x v)))                        -- (5)
    (hB_smooth : ∀ i, ContDiff ℝ 2
      (periodicLift (fun x => B x i)))                        -- (6)
    (hSchwartz : UniformSchwartzDecay f)                      -- (7)
    (hGradBound : ∃ (Cg : ℝ) (Kg : ℕ),
      ∀ (x : Torus3) (v : Fin 3 → ℝ) (i : Fin 3),
      |fderiv ℝ (f x) v (Pi.single i 1)|
        ≤ Cg * (1 + ‖v‖) ^ Kg * f x v)                      -- (8)
    (hVlasov : ∀ x v,
      dotProduct v (torusGradX (fun y => f y v) x) +
      dotProduct (E x + cross v (B x)) (vGrad (f x) v) =
      ν * LandauOperator coulombKernel (f x) v)               -- (9)
    (hAmpere : ∀ x,
      torusCurlX B x = fun i => ∫ v, v i * f x v)           -- (10)
    (hGauss : ∀ x,
      torusDivX E x = (∫ v, f x v) - ρ_ion)                 -- (11)
    (hDivB : ∀ x, torusDivX B x = 0)                        -- (12)
    :
    ∃ (T_eq : ℝ) (B₀ : Fin 3 → ℝ), 0 < T_eq ∧
    (∀ x v, f x v = equilibriumMaxwellian ρ_ion T_eq v) ∧
    (∀ x, E x = 0) ∧
    (∀ x, B x = B₀)
\end{lstlisting}
\caption{The formal \lean~4 statement of the main theorem (\texttt{CoulombConcreteTheorem42}). The proof (omitted) is 80 lines of tactic invocations that assemble the seven-step argument described in \Cref{sec:proof-arch}. Here \texttt{Torus3} is \texttt{Fin\,3\,$\to$\,AddCircle\,1}, \texttt{periodicLift} lifts torus functions to $\RR^3$ for differentiation, and \texttt{equilibriumMaxwellian} is the Maxwellian $\rho_{\mathrm{ion}}/(2\pi T)^{3/2}\exp(-|v|^2/2T)$.}
\label{fig:lean-statement}
\end{figure}

\subsection{Proof Architecture}
\label{sec:proof-arch}

\begin{figure}[t]
  \centering
  \includegraphics[width=\textwidth]{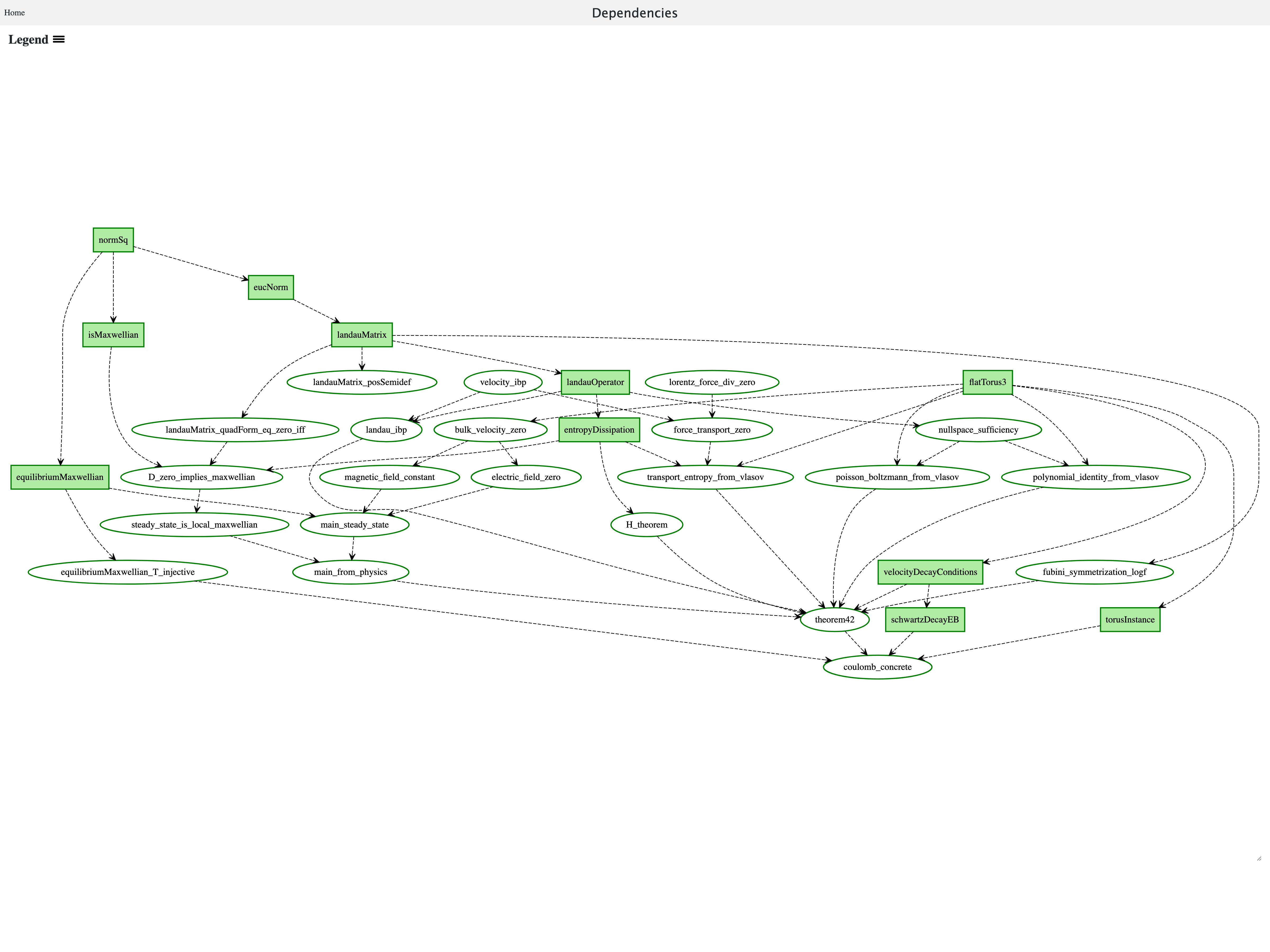}
  \caption{Proof dependency graph, generated with LeanBlueprint \citep{leanblueprint}. All nodes are green (fully proved). The graph flows from primitive definitions (\texttt{normSq}, \texttt{eucNorm}, \texttt{landauMatrix}) at the top through intermediate results (\texttt{H\_theorem}, \texttt{D\_zero\_implies\_maxwellian}, the field and conservation lemmas) down to \texttt{theorem42} and its concrete Coulomb instantiation \texttt{coulomb\_concrete} at the bottom. An interactive version is available in the repository.}
  \label{fig:dep-graph}
\end{figure}

The proof follows a classical entropy method in seven steps (\Cref{fig:dep-graph}):
\begin{enumerate}[leftmargin=2em]
  \item \textbf{H-theorem:} Entropy dissipation $D(f) \leq 0$; the Landau matrix is positive semi-definite.
  \item \textbf{Nullspace:} $D(f) = 0$ implies $f$ is a local Maxwellian: $\log f(x,v) = a(x) + b(x) \cdot v + c(x)|v|^2$.
  \item \textbf{Transport:} Substituting into the Vlasov equation yields a polynomial in $v$ that must vanish identically.
  \item \textbf{Polynomial matching:} The $O(|v|^3)$ terms force $\nabla c = 0$ (constant temperature); the $O(|v|^2)$ terms yield Killing's equation for $b$.
  \item \textbf{Killing's equation:} On the flat torus, the only smooth Killing fields are constants, so $b = b_0$.
  \item \textbf{Maximum principle:} Amp\`ere's law forces $b_0 = 0$; Gauss's law forces $E = 0$ and uniform density.
  \item \textbf{Harmonic analysis:} With $J = 0$, curl $B = 0$ and div $B = 0$ imply $B$ is harmonic on $\TT^3$, hence constant.
\end{enumerate}

\section{Tools}
\label{sec:tools}

The project used four AI systems, none of which required any \lean\ code to be written by hand.

\paragraph{Claude Code} (Anthropic). An agentic coding tool that operates in the terminal, with access to file reading/editing, shell commands, and MCP (Model Context Protocol) integrations. Claude Code was the primary workhorse: it generated all \lean\ code, managed the project structure, ran diagnostics via the \lean\ LSP, and orchestrated submissions to Aristotle. The subscription costs \$200/month (Claude Max plan).

\paragraph{Gemini DeepThink} (Google DeepMind). A mathematical reasoning model accessed via MCP. Gemini generated the initial natural-language proof of the theorem in a five-turn dialogue, identified the relevant literature, and was consulted during the project for planning decisions (e.g., choosing the torus representation).

\paragraph{Aristotle} (Harmonic). A cloud-based automated theorem prover for \lean~4 \citep{aristotle2025}. Lemma statements were extracted from the project, submitted via a Python script, and results were automatically integrated. Aristotle proved 111 lemmas, disproved 28 false conjectures, and returned 66 with \sorry's still present. The service was free during the project period.

\paragraph{Lean LSP tools.} The \lean\ Language Server Protocol, accessed via MCP, provided real-time compilation feedback, goal-state inspection, tactic suggestions, and library search (LeanSearch, Loogle). These tools generated 3{,}194 calls over the project.

\section{Process}
\label{sec:process}

\subsection{Timeline}

The project unfolded over 10 days of active development (March 1 -- 10, 2026), preceded by the Gemini DeepThink session on February 28. \Cref{tab:timeline} summarizes the key milestones.

\begin{table}[t]
\centering
\small
\begin{tabular}{@{}lll@{}}
\toprule
Date & Event & Sorry count \\
\midrule
Feb 28 & Gemini generates natural-language proof & -- \\
Mar 1 & First \lean\ code committed (monolithic file) & $\sim$25 \\
Mar 2 & First Aristotle submissions; hypothesis creep begins & 25 \\
Mar 3 & File split into \texttt{main/} directory; Aristotle results arrive & $25 \to 7$ \\
Mar 5 & Concrete torus representation chosen (with Gemini) & $7 + 22$ \\
Mar 7 & Abstract theorem reaches 0 \sorry's & 0 \\
Mar 7 & \texttt{/critique} adversarial review introduced & 0 \\
Mar 8 & \babysit\ automation loop created (1:09 AM) & 0 \\
Mar 8 & Coulomb kernel instantiation begun; university server online & $+35$ \\
Mar 9 & Coulomb theorem reaches 0 \sorry's (11:15 PM) & 0 \\
Mar 10 & Cleanup ($-3$K lines); non-vacuousness theorem & 0 \\
\bottomrule
\end{tabular}
\caption{Project timeline and sorry count evolution.}
\label{tab:timeline}
\end{table}

\subsection{Phase 1: Natural-Language Proof (Feb 28)}

The project began with a question posed to Gemini DeepThink:

\begin{quote}
\emph{``What is known about the equilibrium of the Vlasov-Maxwell-Landau equations with Coulomb collisions, and a single species? Does it have to be a global Maxwellian? With what covariance? How is it shown?''}
\end{quote}

Gemini produced a detailed six-step proof, identified the relevant literature \citep{villani2002,cercignani1994,guo2003,guo2012,desvillettes2005}, and refined the argument over five turns. Crucially, the proof required no mathematical corrections: the mathematician verified the argument by hand and found it complete and correct, needing only the identification of exact supporting theorems from the cited papers. The final natural-language proof became the blueprint for the formalization.

\subsection{Phase 2: Scaffolding (Mar 1 -- 2)}

The formalization prompt was:

\begin{quote}
\emph{``Look at H-theorem-formal.tex. I want to formalize that the steady state of the VML.''}
\end{quote}

Claude Code generated a monolithic 1{,}000+ line \lean\ file within hours, with \sorry's marking every gap. Claude estimated the full formalization would require ``multiple months.'' The mathematician's role was to reject false paths and enforce design principles, particularly hypothesis discipline:

\begin{quote}
\emph{``You keep doing the same thing -- adding hypotheses that are actually lemmas to be proven! Why do you do that?''} (Mar 2)
\end{quote}

This became the project's central methodological principle, codified in \texttt{CLAUDE.md}: a \sorry\ (an acknowledged gap) is preferable to an unnecessary hypothesis (which silently weakens the theorem).

\subsection{Phase 3: Abstract Proof Chain (Mar 3 -- 7)}

The monolithic file was split into a \texttt{main/} directory with separate files for each section of the proof. A \texttt{FlatTorus3} typeclass abstracted the spatial domain, specifying integration by parts, curl/divergence identities, a maximum principle, and constancy of harmonic functions -- without fixing a particular manifold. The mathematical argument (Sections 2 -- 8) was formalized against this interface.

A concrete \texttt{TorusInstance} module proved that $\TT^3 = (\RR/\mathbb{Z})^3$ satisfies all 22 fields of \texttt{FlatTorus3}, using the \mathlib\ representation \texttt{Fin\,3\,$\to$\,AddCircle\,1}. The representation was chosen in consultation with Gemini. Periodicity is automatic: functions on the torus factor through the quotient map $\RR \to \RR/\mathbb{Z}$.

By March 7, all \sorry's in the abstract proof chain were closed -- the abstract theorem was fully proved.

\subsection{Phase 4: Automation (Mar 7 -- 8)}

On March 7, an adversarial self-review process was introduced:

\begin{quote}
\emph{``You are a hostile reviewer trying to REJECT this formalization. Your job is to find every weakness, gap, and dishonesty.''}
\end{quote}

The output, \texttt{critique.md}, identified false hypotheses, unnecessary assumptions, and dead code. On March 8 at 1:09 AM, the automation suite was created in a 20-minute session. The initial \babysit\ loop combined four slash commands (\texttt{/critique}, \texttt{/prove}, \texttt{/check-aristotle}, \texttt{/cleanup}) and was set to run on a 10-minute timer. Over the following days, the loop was iteratively expanded to 11 steps -- adding \texttt{/plan}, \texttt{/simplify}, \texttt{/strengthen}, \texttt{/submit-aristotle}, \texttt{/log}, \texttt{/commit}, and \texttt{/alert} (Telegram notifications) -- and executed 122 documented cycles over the project.

\subsection{Phase 5: Coulomb Kernel (Mar 8 -- 9)}

With the abstract theorem proved, the final challenge was instantiating it for the physical Coulomb kernel $\Psi(r) = r^{-3}$. This required proving that the Landau collision operator satisfies all 17 integrability, differentiability, and continuity conditions in the \texttt{VelocityDecayConditions} bundle.

The Coulomb singularity at $r = 0$ makes $|A_{ij}(z)| \sim \|z\|^{-1}$, which is barely integrable in $\RR^3$. Each integrability proof required decomposing the domain into a ball around the origin (handled by local integrability of $\|z\|^{-1}$) and the complement (handled by Schwartz decay). This phase was the hardest part of the formalization, producing $\sim$4{,}000 lines of analytical estimates.

A university server was brought online on March 8 to run \babysit\ loops concurrently with the local machine, enabling continuous progress. The sorry count dropped from 35 to 0 on March 9, with the last sorry closed at 11:15 PM.

\subsection{Phase 6: Cleanup (Mar 10)}

With zero \sorry's achieved, the \babysit\ loop shifted to code quality: removing dead code ($-3$K lines), eliminating unnecessary \texttt{maxHeartbeats} overrides, and re-enabling linters. A separate non-vacuousness theorem (\texttt{CoulombConcreteTheorem42\_nonvacuous}) verified that the 12 hypotheses are satisfiable -- the equilibrium Maxwellian with $E = 0$, $B = 0$ witnesses all conditions. This sanity check was suggested independently by expert reviewers on the Lean Zulip.

\section{Statistics}
\label{sec:stats}

\begin{table}[t]
\centering
\begin{tabular}{@{}lr@{}}
\toprule
Metric & Value \\
\midrule
\lean~4 files & 34 \\
Lines of code & 10{,}445 \\
Theorems & 39 \\
Lemmas & 186 \\
Definitions & 28 \\
\sorry's & 0 \\
Development period & 10 days \\
Git commits & 213 \\
Claude Code sessions & 32 (across 2 machines) \\
Human prompts (excl.\ slash commands) & 229 \\
\babysit\ cycles & 122 \\
Assistant turns & 27{,}186 \\
Tool calls & 17{,}334 \\
Tokens consumed (input) & 2.8 billion \\
Tokens consumed (output) & 11 million \\
Input-to-output ratio & 254:1 \\
Aristotle submissions & 220 \\
\quad Proved & 111 (50\%) \\
\quad Disproved & 28 (13\%) \\
\quad Returned with \sorry & 66 (30\%) \\
\quad Failed & 15 (7\%) \\
\bottomrule
\end{tabular}
\caption{Project statistics. The 229 human prompts exclude bare slash commands like \babysit; all are archived with timestamps. The mathematician estimates $\sim$50 hours of active supervision over 10 days.}
\label{tab:stats}
\end{table}

\Cref{tab:stats} summarizes the project. The 2.8 billion input tokens are dominated by cache reads of the \lean\ codebase and \mathlib\ context -- the 254:1 input-to-output ratio reflects the read-heavy nature of formal verification. At Opus~4 API pricing with prompt caching, this would cost approximately \$6{,}300: \$4{,}071 in cache reads (\$1.50/M), \$1{,}401 in cache creation (\$18.75/M), \$826 in output (\$75/M), and \$6 in fresh input (\$15/M). Without prompt caching, the same workload would cost $\sim$\$42{,}700 -- caching saves 85\%. The mathematician paid only \$200 for the Claude Max subscription, which provides generous usage.

\begin{figure}[t]
  \centering
  \includegraphics[width=\textwidth]{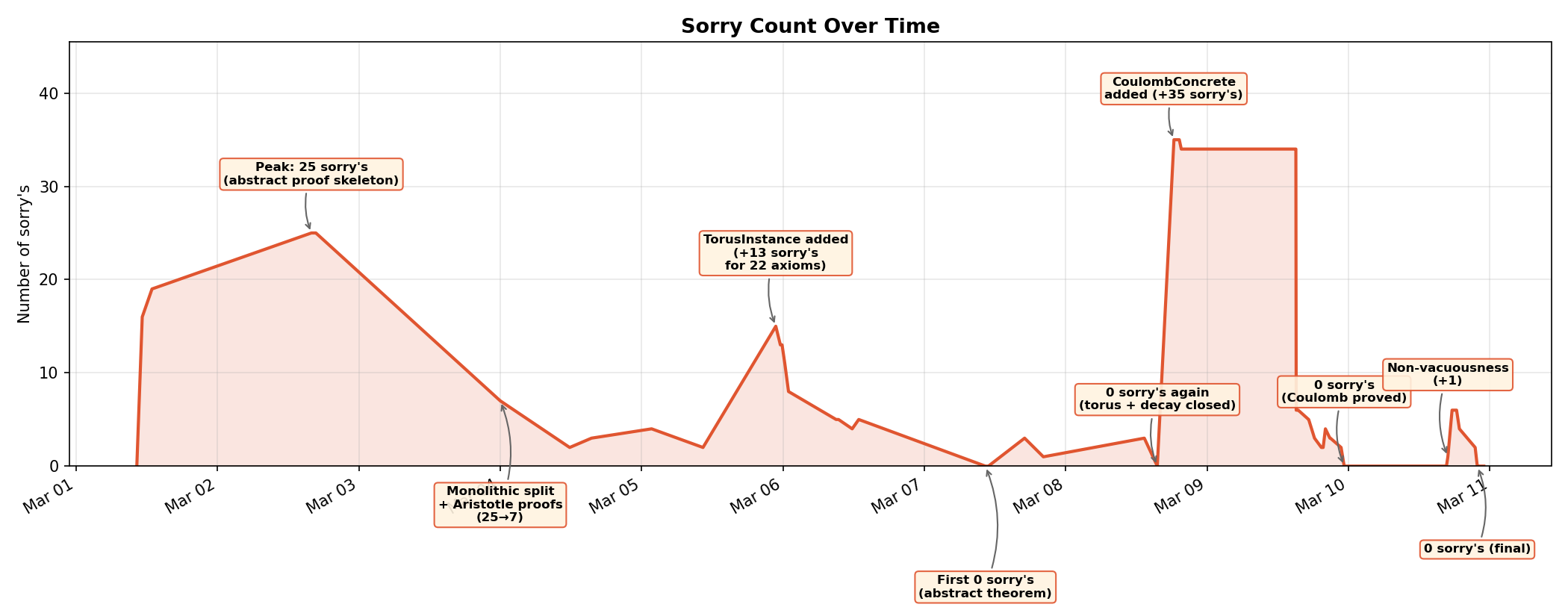}
  \caption{Sorry count in \texttt{main/} over time (85 commits). The sawtooth pattern reflects alternating phases of scaffolding (sorry's accumulate) and proving campaigns (sorry's eliminated). Key events: peak 25 on Mar~2 (abstract proof chain stated); Aristotle drops count from 25$\to$7 on Mar~3; first 0 sorry's on Mar~7 (abstract theorem proved); $+$35 sorry's on Mar~8 (Coulomb kernel instantiation begun); 35$\to$0 sprint on Mar~9; brief $+$1 on Mar~10 (non-vacuousness theorem, immediately resolved).}
  \label{fig:sorry}
\end{figure}

\begin{figure}[t]
  \centering
  \includegraphics[width=\textwidth]{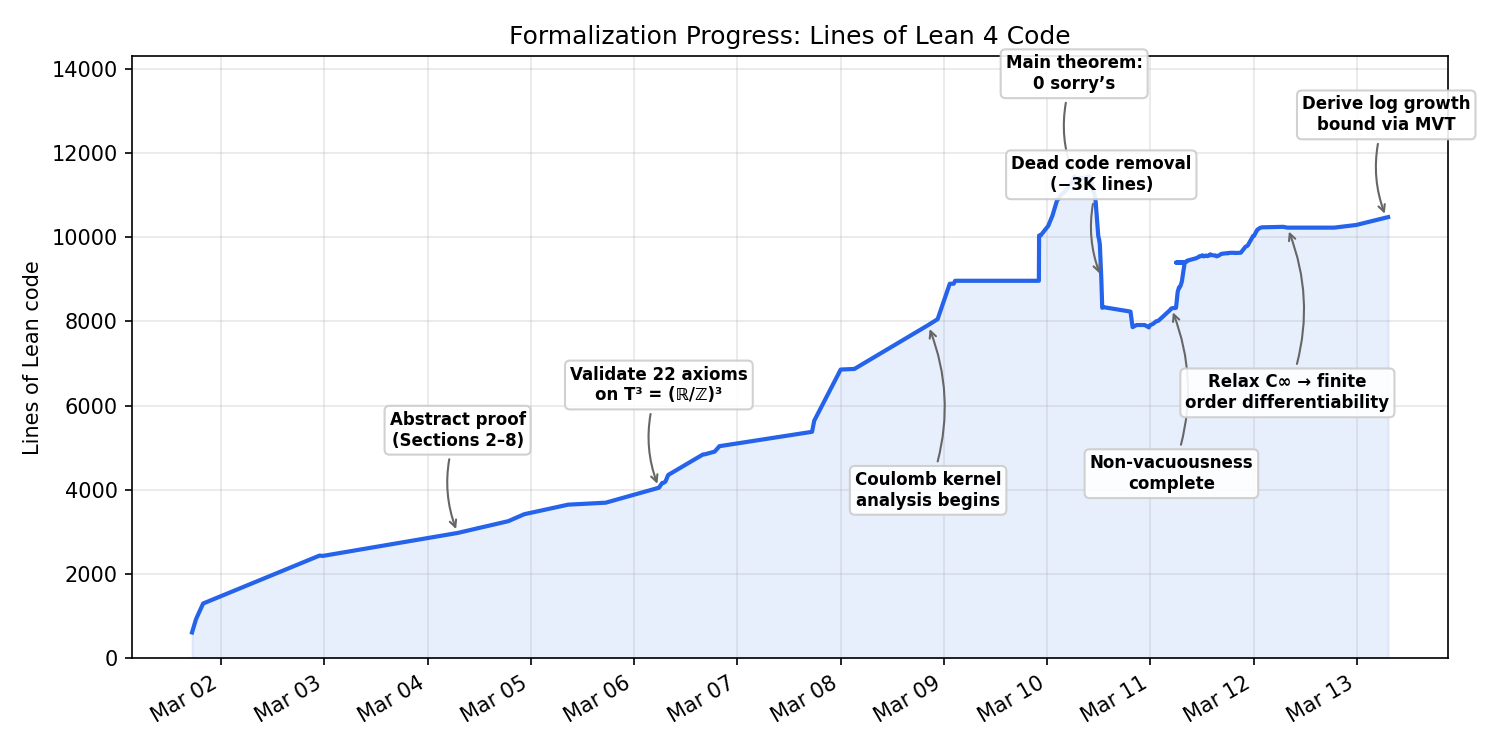}
  \caption{\lean\ lines of code over time (all 213 commits). The project progressed in four phases: (1) Abstract proof chain (Mar~3 -- 6), formalized against an abstract \texttt{FlatTorus3} typeclass. (2) Coulomb kernel analysis (Mar~7 -- 9), handling the singularity at $r=0$ -- the sharp LOC increase reflects $\sim$4K lines of analytical estimates. (3) Cleanup (Mar~10, early): $\sim$3K lines of dead code, redundant lemmas, and unnecessary heartbeat overrides removed. (4) Non-vacuousness (Mar~10, late): satisfiability of the 12 hypotheses.}
  \label{fig:loc}
\end{figure}

\begin{figure}[t]
  \centering
  \includegraphics[width=\textwidth]{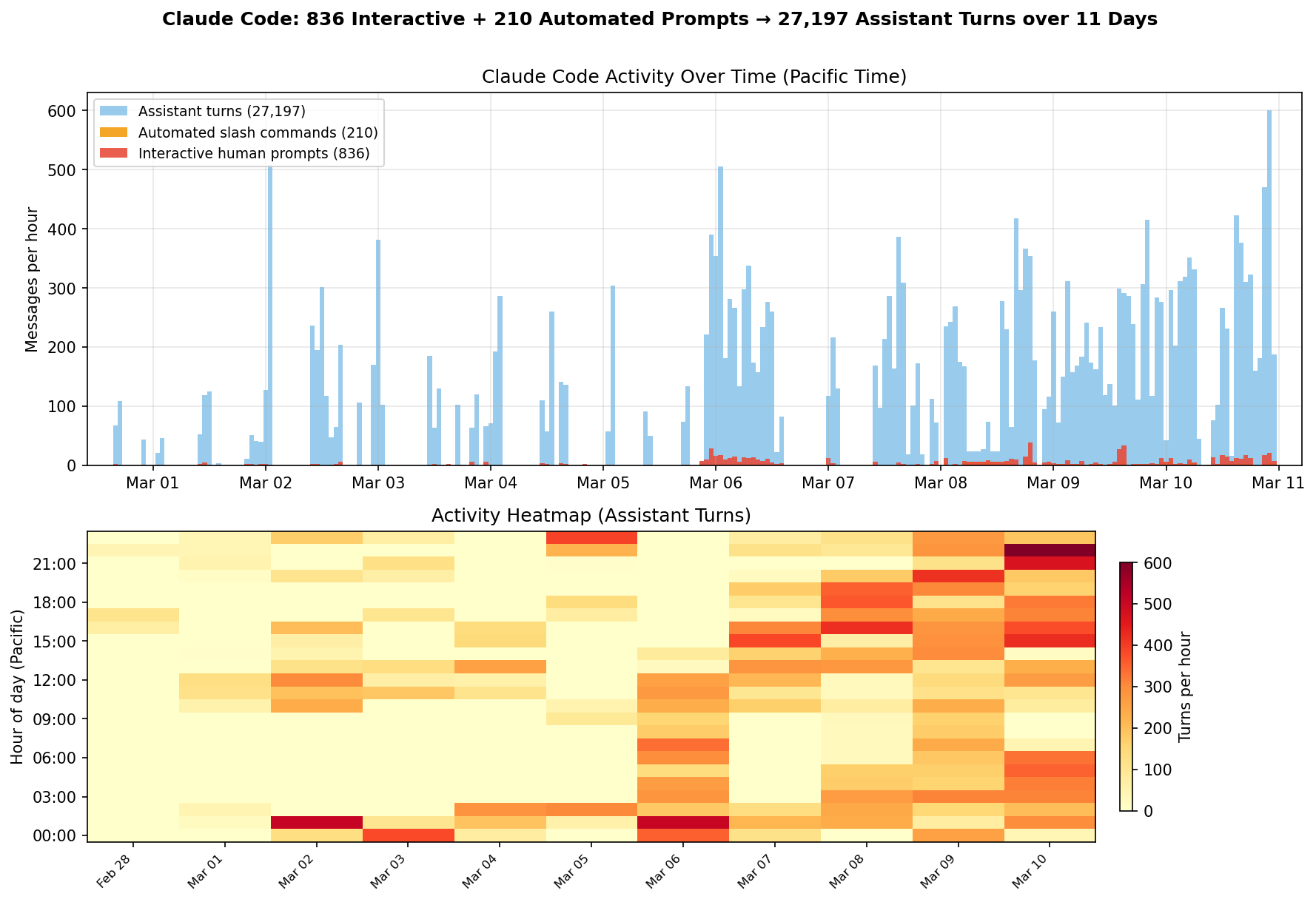}
  \caption{Claude Code activity across two machines (local $+$ university server). Top: messages per hour (red = interactive human, orange = automated, blue = assistant). Bottom: hour-of-day heatmap. The 27{,}186 assistant turns versus 229 human prompts give a 119:1 ratio, reflecting the high autonomy of each \babysit\ cycle (typically 20 -- 50 tool calls per invocation). The heatmap shows near-round-the-clock activity on Mar~9 -- 10 when both machines ran \babysit\ loops concurrently.}
  \label{fig:activity}
\end{figure}

\begin{figure}[t]
  \centering
  \includegraphics[width=\textwidth]{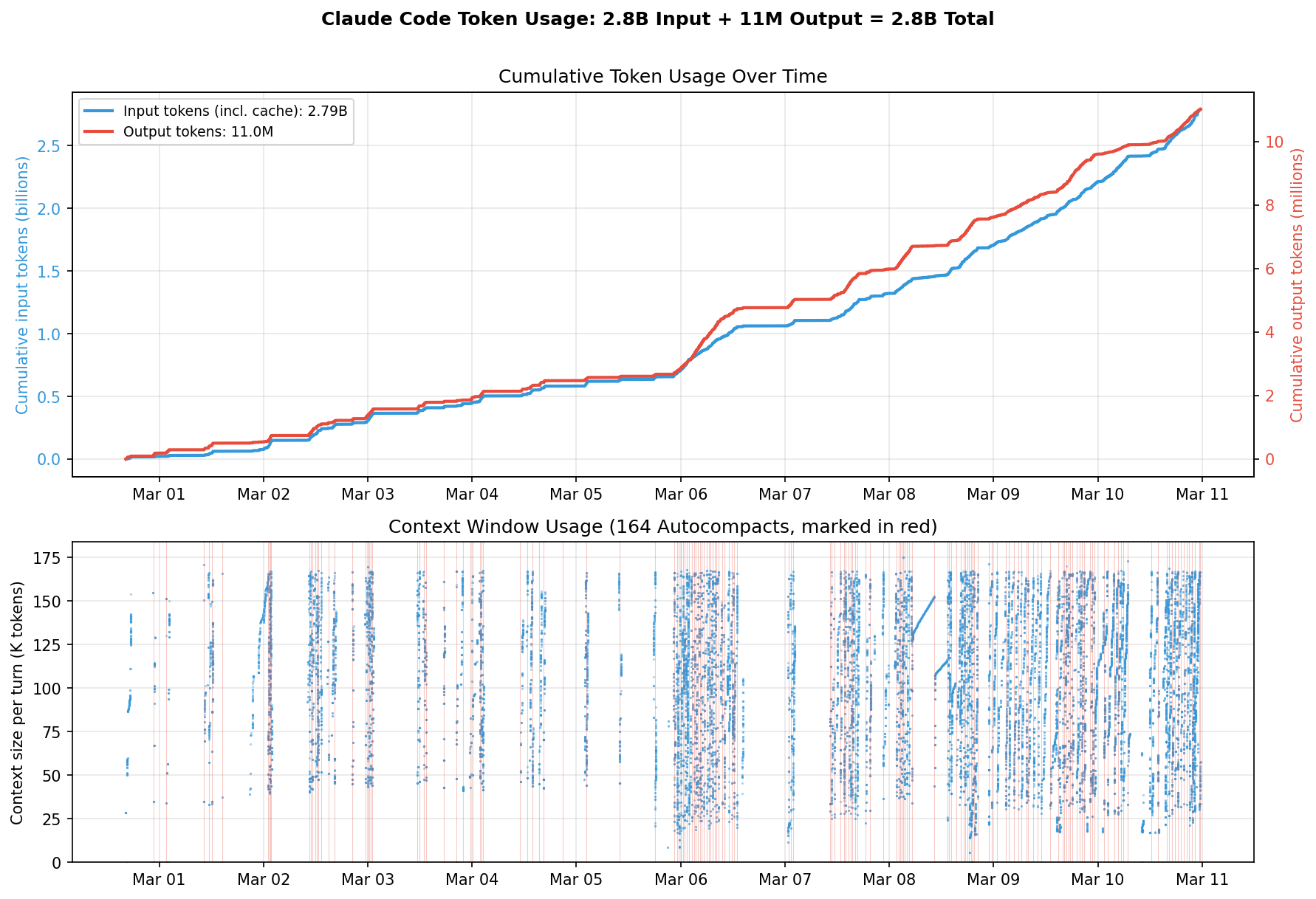}
  \caption{Token consumption across the project. Top: cumulative input tokens (blue, 2.8B) and output tokens (red, 11M). The 254:1 input-to-output ratio reflects the read-heavy nature of formal verification -- Claude reads far more context than it writes. Bottom: per-turn context size, showing a sawtooth pattern as the context window fills to $\sim$175K tokens and resets at each of the 164 autocompacts (red vertical lines). Dense compaction clusters on Mar~6 -- 10 correspond to the intensive Coulomb kernel analysis and \babysit\ loop periods.}
  \label{fig:tokens}
\end{figure}

\begin{figure}[t]
  \centering
  \includegraphics[width=\textwidth]{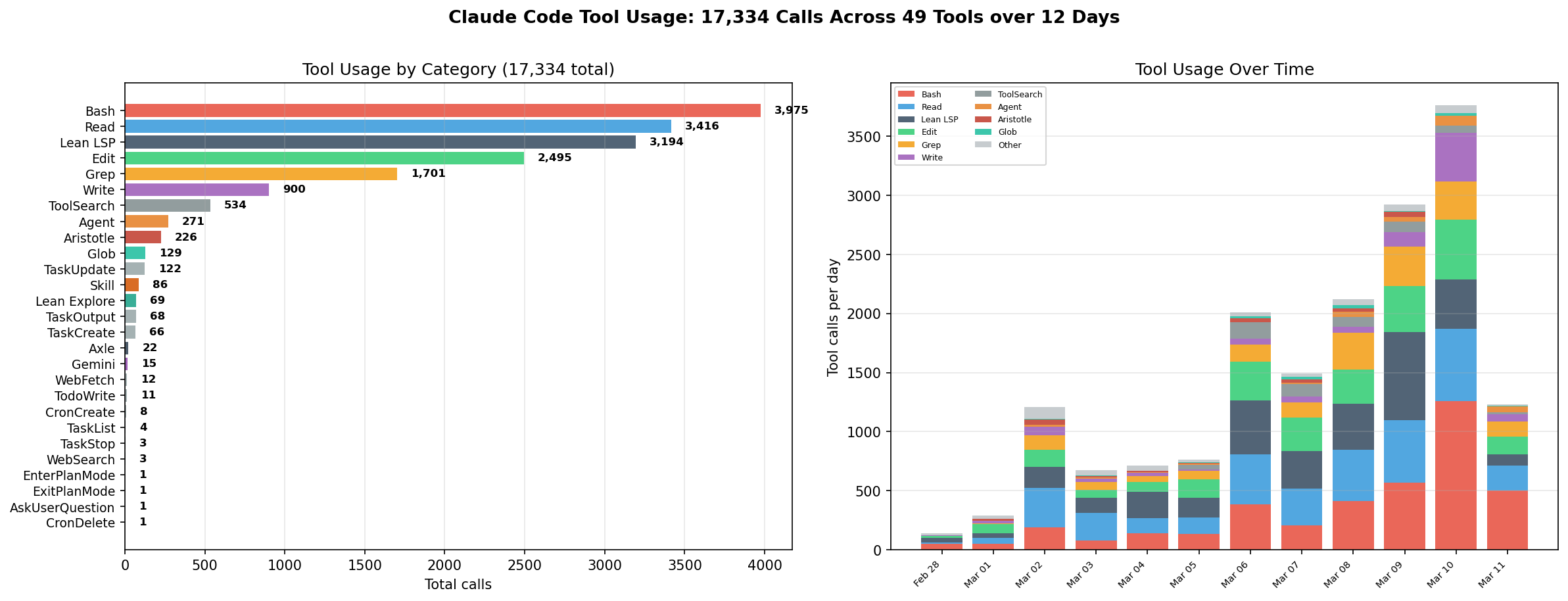}
  \caption{Tool usage across 17{,}334 invocations. Left: totals by category. The core editing loop dominates -- Bash (3{,}975), Read (3{,}416), Edit (2{,}495), Grep (1{,}701), Write (900) -- reflecting the iterative nature of formal verification. \lean\ LSP tools (3{,}194 combined) provided real-time feedback. Right: daily breakdown showing the ramp-up from exploratory work (Feb~28 -- Mar~2) through peak activity (Mar~6 -- 10) when both machines ran \babysit\ loops concurrently.}
  \label{fig:tools}
\end{figure}

\begin{figure}[t]
  \centering
  \includegraphics[width=\textwidth]{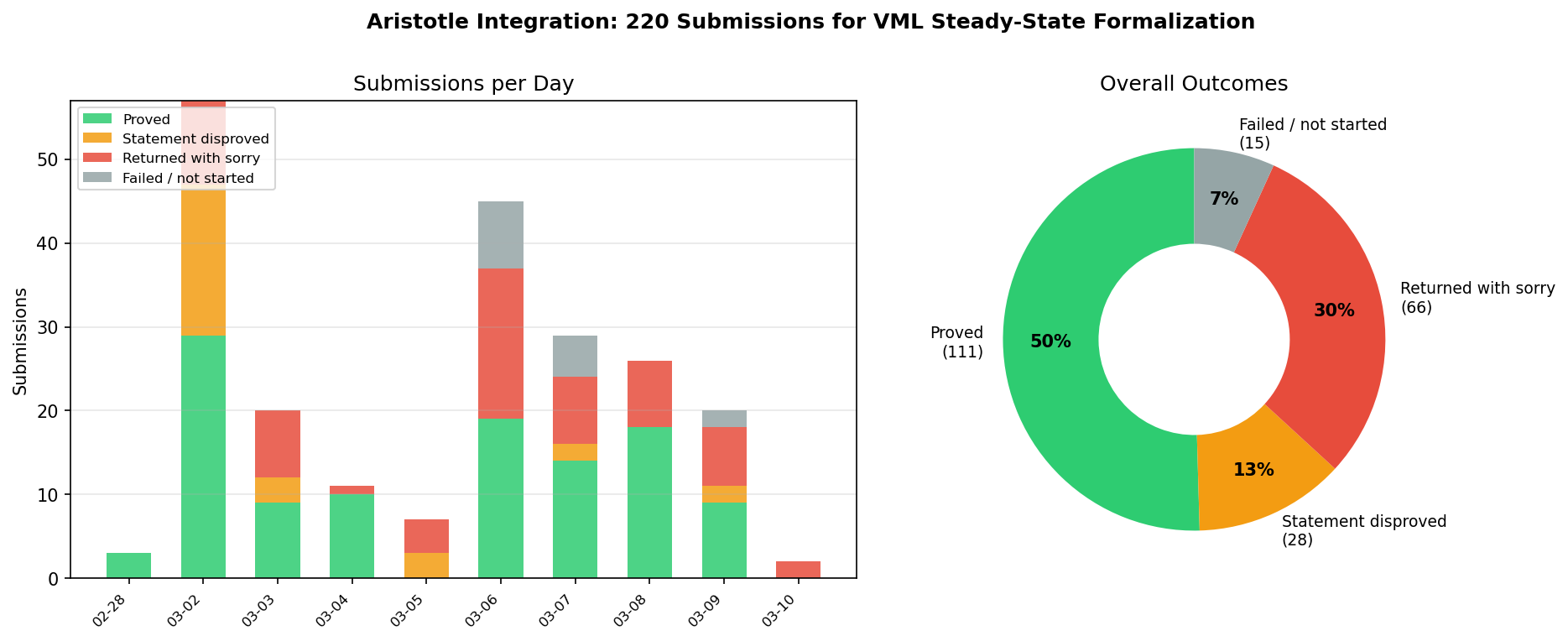}
  \caption{Outcomes of 220 Aristotle submissions. Aristotle proved 111 lemmas cleanly (50\%), disproved 28 false statements (13\%), returned 66 with sorry's still present (30\%), and 15 failed or were never started (7\%). The disproved statements were particularly valuable -- they caught false conjectures early (e.g., missing measurability hypotheses, Vitali-set counterexamples, incorrect Schwartz decay estimates for the Coulomb kernel). The stacked bar chart shows peak submission activity on Mar~2 (57 jobs) and Mar~6 (45 jobs).}
  \label{fig:aristotle}
\end{figure}

\begin{figure}[t]
  \centering
  \includegraphics[width=\textwidth]{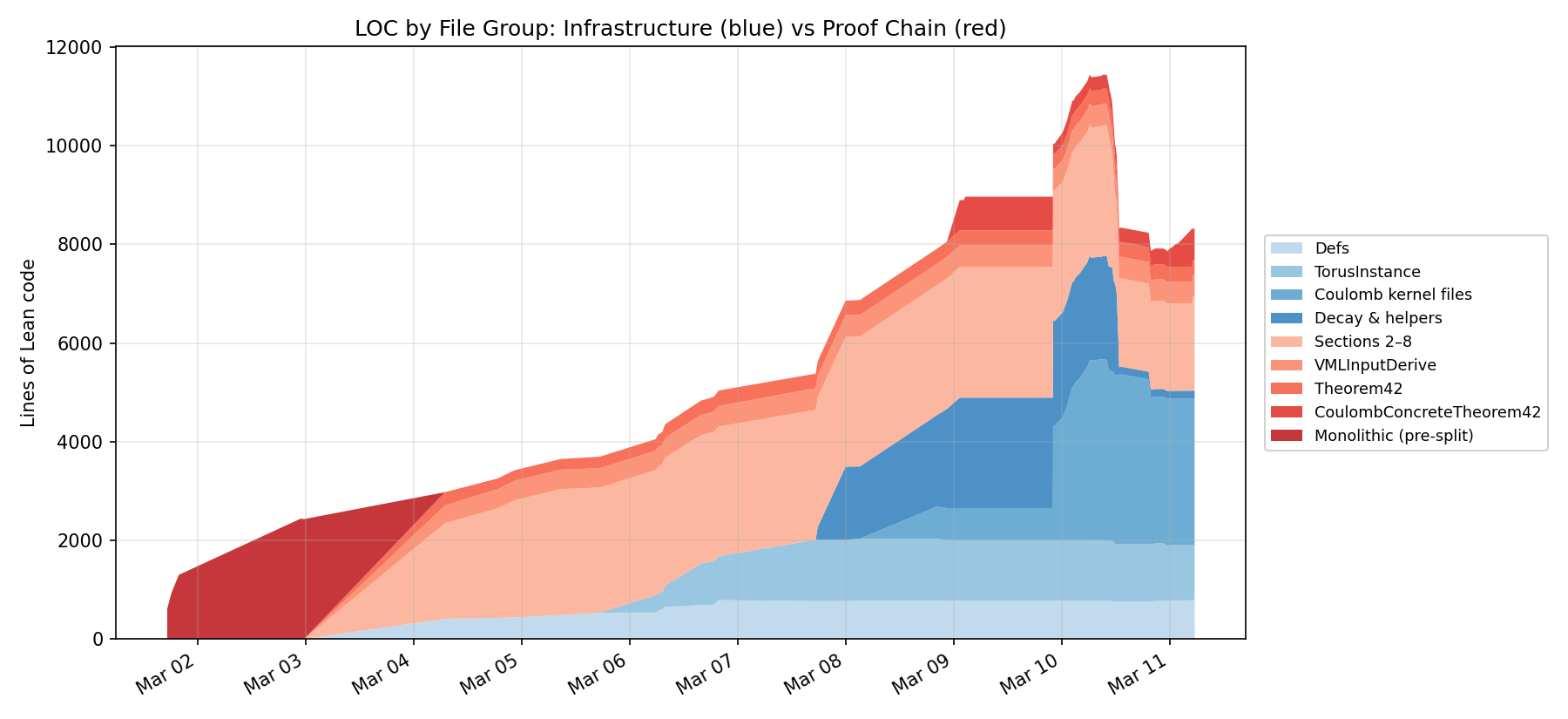}
  \caption{LOC by file group over time. Red shades are the mathematical proof chain (Sections~2--8, \texttt{Theorem42}, \texttt{VMLInputDerive}, \texttt{CoulombConcreteTheorem42}); blue shades are supporting infrastructure (definitions, torus instance, Coulomb kernel integrability files, decay helpers). At completion, infrastructure accounts for roughly half of the codebase -- reflecting the substantial gap between \mathlib's current coverage and the background material needed for research-level mathematical physics. The sharp blue growth on Mar~7--8 is the Coulomb kernel analysis ($\sim$4K lines of analytical estimates for the singularity at $r=0$). The cleanup phase (Mar~10) primarily removed blue infrastructure code, while the red proof chain remained stable.}
  \label{fig:loc-breakdown}
\end{figure}

The development trajectory is shown in \Cref{fig:sorry,fig:loc}: the sorry count follows a sawtooth pattern as scaffolding and proving alternate, while the LOC history shows the four-phase structure of the project. \Cref{fig:loc-breakdown} reveals that roughly half of the final codebase is supporting infrastructure -- definitions, the torus instance, Coulomb kernel integrability files, and decay helpers -- rather than the core proof chain. This reflects the substantial gap between \mathlib's current coverage and the background material needed for research-level mathematical physics: the Landau operator, torus differential calculus, Schwartz decay machinery, and Coulomb singularity estimates all had to be built from scratch. \Cref{fig:activity} shows session activity across two machines, with near-round-the-clock coverage during the final sprint. \Cref{fig:tokens} breaks down token consumption, revealing the 254:1 input-to-output ratio characteristic of formal verification. \Cref{fig:tools} summarizes tool usage across 17{,}334 invocations, and \Cref{fig:aristotle} shows the outcomes of 220 Aristotle submissions.

\begin{figure}[t]
  \centering
  \includegraphics[width=\textwidth]{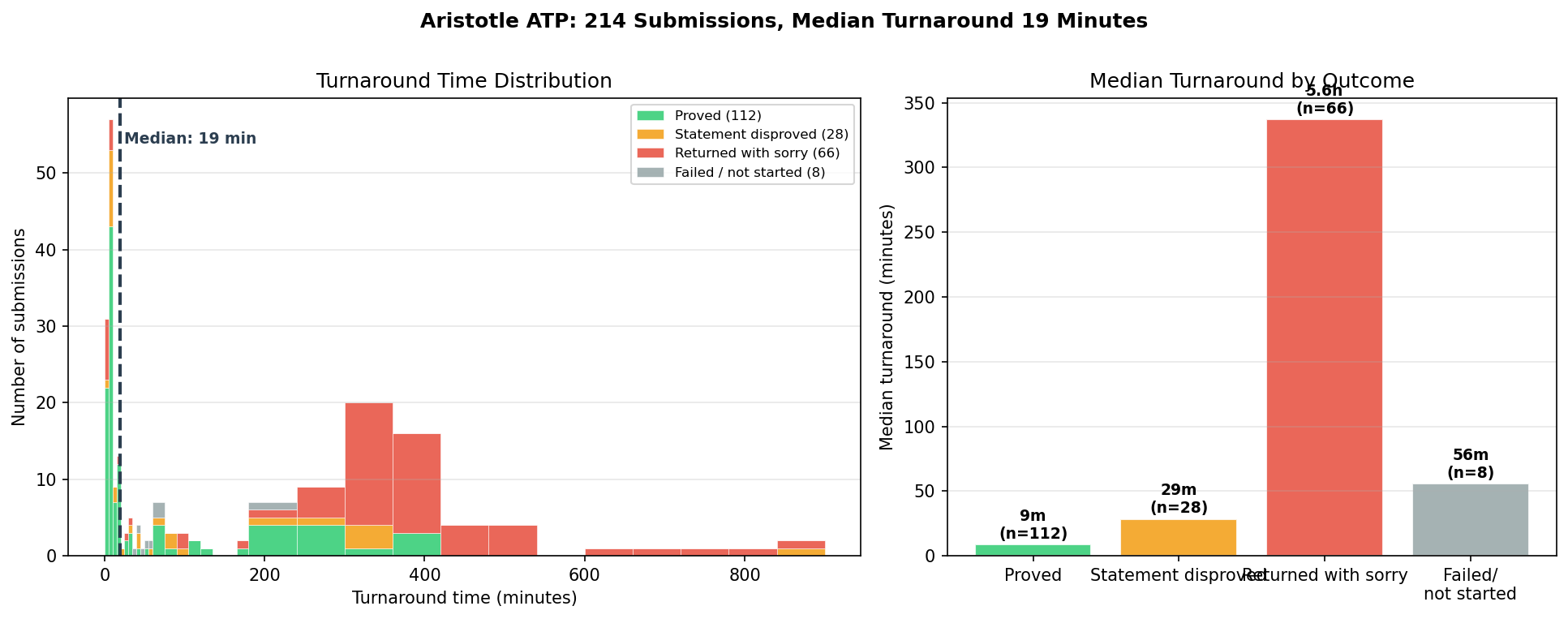}
  \caption{Aristotle turnaround times for 214 submissions, colored by outcome. Left: histogram. Proved lemmas (green) cluster in the first bins (median 9 minutes), while ``returned with sorry'' (red) dominates the long tail. Right: median turnaround by category. The overall median was 19 minutes. Disproved statements took 29 minutes (counterexample construction); submissions returned with sorry took a median of 5.6 hours -- Aristotle exhausts its time budget before giving up.}
  \label{fig:aristotle-time}
\end{figure}

\begin{figure}[t]
  \centering
  \includegraphics[width=\textwidth]{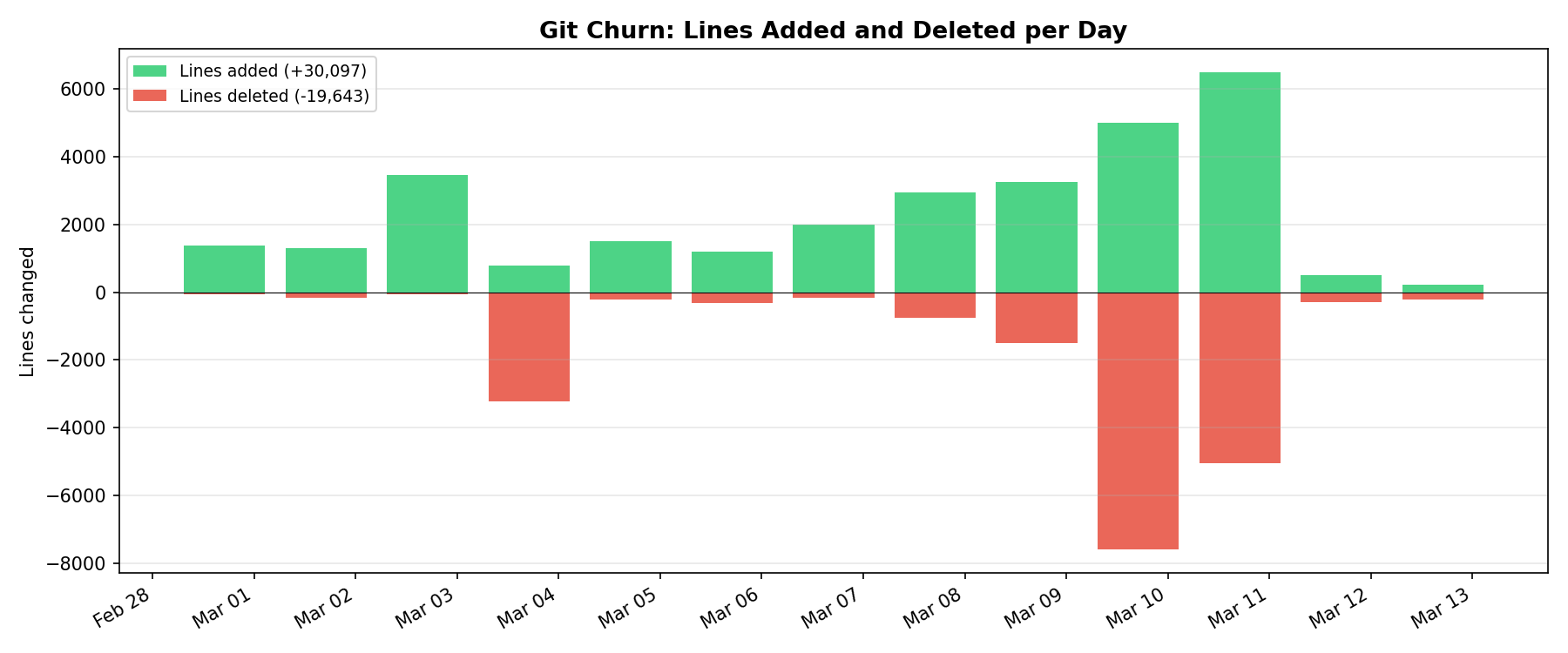}
  \caption{Lines of \lean\ code added and deleted per day. Most days are net-positive (new content), with two exceptions: Mar~3 (monolithic file split into \texttt{main/} -- same code reorganized) and Mar~9 -- 10 (cleanup phase removed $\sim$3K lines of dead code, redundant lemmas, and unnecessary heartbeat overrides). Total: $+$21{,}783 added, $-$13{,}478 deleted, net $+$8{,}305 lines.}
  \label{fig:churn}
\end{figure}

\section{Lessons Learned}
\label{sec:lessons}

\subsection{Hypothesis Discipline is Critical}

The most persistent failure mode of Claude Code was \emph{hypothesis creep}: when faced with a difficult lemma, the agent's default behavior was to assume it as a hypothesis of the main theorem rather than prove it. By March 4, the main theorem had accumulated 42 hypotheses -- most of which should have been derived lemmas.

This failure mode is insidious because it is syntactically valid: the \lean\ code compiles, the sorry count stays low, and the proof ``works.'' But the theorem statement is progressively weakened. In the extreme case, one could ``prove'' any theorem by assuming its conclusion as a hypothesis.

The fix required persistent human intervention:

\begin{quote}
\emph{``I have told you time and again to keep the hypotheses to Theorem42 minimal. If you need a mathematical fact, it should be a lemma with a sorry, not a hypothesis!''} (Mar 3)
\end{quote}

This was codified in \texttt{CLAUDE.md}: \emph{``The goal is not to end up with 0 sorry's! The goal is to make an honest formalization of the main theorem, with only the genuinely needed mathematical/physical assumptions.''} The 42 hypotheses were systematically reduced: velocity-space decay conditions were bundled into a single \texttt{VelocityDecayConditions} structure (replacing 15 scattered hypotheses), and regularity facts like continuity of the density $\rho$ were derived inside the proof from domination bounds. The final concrete theorem has 12 hypotheses, all physically meaningful.

\subsection{Aristotle is Transformative}

The cloud-based theorem prover Aristotle \citep{aristotle2025} was the single most impactful tool after Claude Code itself. Of 220 submissions, 111 were proved outright (50\%), and 28 were \emph{disproved} -- Aristotle found counterexamples or proved the negation.

The disproved statements were arguably more valuable than the proved ones. They caught false conjectures early: missing measurability hypotheses, claims that fail on pathological counterexamples (e.g., Vitali sets), and Schwartz decay estimates that are actually false for the Coulomb kernel. The fix-and-resubmit cycle served as the primary mechanism for catching bugs in the formalization.

Aristotle's overall median turnaround was 19 minutes (\Cref{fig:aristotle-time}). Proved lemmas returned in a median of 9 minutes; disproved statements took 29 minutes (counterexample construction); but submissions returned with \sorry\ had a median of 5.6 hours, as Aristotle exhausted its time budget searching before giving up. We recommend decomposing proofs aggressively into small lemmas and submitting each to an automated prover -- even lemmas that seem hard.

\subsection{Definition Alignment Requires Expert Review}

After the formalization was announced on the Lean Zulip, S\'ebastien Gou\"ezel (a Mathlib maintainer) immediately identified a definition-alignment bug: the original theorem used \texttt{ContDiff $\RR$ $\top$}, which in \mathlib\ means \emph{analytic}, not $C^\infty$ as intended. Neither the AI agent nor the mathematician caught this during 10 days of development. Gou\"ezel remarked: \emph{``I find it a little bit worrisome that AI can overlook such discrepancies -- if there is one in the only line I've checked, how many others are there?''}

The fix was straightforward -- replacing $\top$ with explicit finite smoothness ($C^3$ in velocity, $C^2$ in space) -- and the resulting theorem is actually \emph{stronger}, since it assumes less. But the episode illustrates a fundamental risk of AI-generated formalizations: the code compiles and the proof checks, yet the theorem does not state what was intended. Gou\"ezel also suggested formalizing that the equilibrium Maxwellian satisfies all 12 hypotheses (the non-vacuousness theorem), which was implemented as an additional sanity check.

This lesson is perhaps the most important for the community: \textbf{compilation is not correctness}. Expert review of definitions and theorem statements remains essential, and cannot yet be delegated to the AI.

\subsection{Automation Helps but is Not Strictly Necessary}

The \babysit\ loop (\Cref{sec:process}) ran for 122 cycles over the project. It was valuable for overnight progress and for maintaining momentum on tedious tasks (linter cleanup, dead code removal). However, the most important breakthroughs (closing the abstract theorem, proving the Coulomb integrability estimates) happened during interactive sessions with direct human guidance.

The automation was most effective for:
\begin{itemize}[leftmargin=2em]
  \item \textbf{Maintaining code quality:} The adversarial \texttt{/critique} step caught issues that would otherwise accumulate (unused imports, stale comments, false docstrings).
  \item \textbf{Aristotle polling:} Checking for completed Aristotle jobs and integrating results.
  \item \textbf{Overnight progress:} Running unattended on two machines while the mathematician slept or traveled.
\end{itemize}

It was less effective for:
\begin{itemize}[leftmargin=2em]
  \item \textbf{Novel proof strategies:} The agent often got stuck in loops, retrying the same failing approach rather than reconsidering the proof strategy.
  \item \textbf{Architectural decisions:} Choosing representations, structuring typeclasses, and deciding what to abstract required human judgment.
\end{itemize}

\subsection{Gemini for Mathematical Reasoning}

Gemini DeepThink served two distinct roles: (1) generating the initial proof blueprint, and (2) providing mathematical advice during the formalization. For the first role, it was excellent -- the proof it produced (\Cref{sec:process}) served as the complete blueprint for the 10K-line formalization. For the second role, it was useful but not essential: Claude Code's own mathematical reasoning was often sufficient for tactical decisions. In the mathematician's own assessment: \emph{``I did not find Gemini to be that terribly useful [beyond the initial proof]. It was mostly for experimental sake.''}

The key insight is that the full loop of mathematical research -- from conjecture to resolution to detailed proof -- can now be completed by AI with the human acting as verifier rather than creator. Proof generation and proof formalization are different skills, and a model that excels at informal mathematical reasoning (Gemini) can be productively combined with a model that excels at code generation (Claude), with each handling its strength. The \lean\ kernel then provides the final guarantee of correctness that neither model can offer alone.

\subsection{The Agent Gets ``Lazy''}

Beyond hypothesis creep, Claude Code exhibited several avoidance behaviors:
\begin{itemize}[leftmargin=2em]
  \item \textbf{Premature \sorry:} Marking a goal as \sorry\ and moving on, rather than attempting the proof.
  \item \textbf{Excessive heartbeats:} Setting \texttt{maxHeartbeats 800000} to suppress timeouts rather than simplifying the proof.
  \item \textbf{Dead code accumulation:} Leaving unused lemmas, commented-out code, and stale helper definitions rather than cleaning up.
  \item \textbf{Over-engineering:} Creating unnecessary abstraction layers, helper functions, and intermediate definitions when a direct proof would suffice.
  \item \textbf{No-op cycles:} During automated \babysit\ runs, the agent would sometimes complete a cycle with no meaningful changes despite open issues in \texttt{critique.md}, requiring prompt rewrites to enforce progress.
\end{itemize}

These behaviors are not bugs per se -- they are rational strategies for an agent optimizing for compilation success rather than proof quality. The fix is explicit instructions (\texttt{CLAUDE.md}) and persistent human supervision. The adversarial \texttt{/critique} step in the \babysit\ loop partially automated this supervision.

\subsection{The Abstract/Concrete Split Pays Off}

Separating the proof into an abstract framework (\texttt{FlatTorus3} typeclass) and a concrete instantiation (\texttt{TorusInstance}) was an early design decision that paid dividends throughout. When the torus proofs required rewrites -- removing false hypotheses, changing the representation of spatial differentiability -- the abstract proof chain (Sections 2 -- 8) was completely unaffected. This separation of concerns is standard software engineering practice, but it is not the default behavior of the AI agent, which tends to produce monolithic proofs.

\subsection{No Lean Code Required}

The mathematician wrote zero lines of \lean\ code and zero lines of any other programming language. All \lean\ code was generated by Claude Code from natural-language prompts. Of the 229 human prompts sent to Claude Code (excluding bare slash commands and context continuations), most were short directives or confirmations. The mathematician estimates approximately 50 hours of active supervision -- an average of 5 hours per day watching Claude Code work in real time.

The mathematician's contributions were:
\begin{enumerate}[leftmargin=2em]
  \item Posing the conjecture.
  \item Checking the final theorem statement (12 hypotheses, each evaluated for physical correctness).
  \item Enforcing hypothesis discipline and architectural decisions.
  \item Providing mathematical intuition when the agent was stuck.
  \item Designing and iterating on the automation suite.
\end{enumerate}

The mathematician did read key definitions and theorem statements in \lean\ to verify alignment with the intended mathematics, but did not debug proofs or lemma statements directly.

\section{Related Work}
\label{sec:related}

\Cref{tab:comparison} compares recent research-level formalizations with publicly available code.

\paragraph{AI-assisted theorem proving.}
Seed-Prover \citep{seedprover2025} achieves state-of-the-art results on competition mathematics using reinforcement learning with \lean\ feedback. Harmonic's Aristotle \citep{aristotle2025} achieved gold-medal performance on the 2025 IMO and auto-formalized GPT-5.2-generated proofs of open Erd\H{o}s problems in \lean\ \citep{erdos728}. Google DeepMind's Aletheia agent \citep{aletheia2026,erdos_gemini2026} autonomously produced research papers in natural language, including solutions to four open Erd\H{o}s conjectures. AxiomProver autonomously generated formal \lean\ proofs of Fel's open conjecture \citep{axiom2026fel}, a result on partial regularity of primes \citep{axiom2026partial}, and parity of $k$-differentials \citep{axiom2026parity} -- all from natural-language problem statements with zero human guidance. Unlike our work, these systems are specialized for theorem proving, and the formalized results are typically short combinatorial or algebraic arguments rather than large-scale formalizations requiring hard analysis (Coulomb singularity estimates, PDE energy methods).

\paragraph{Agentic formalization.}
Numina-Lean-Agent \citep{numina2025} combines Claude Code with MCP tools and formalized effective Brascamp-Lieb inequalities ($\sim$8K lines) with two human experts collaborating on the \lean\ code. Archon \citep{archon2025} uses a dual-agent architecture and demonstrated full automation of FirstProof problems. \citet{davidbayer2025} formalized their own new theorem on Diophantine complexity bounds in 20{,}000 lines of Isabelle \emph{in parallel} with the mathematical research -- an ``in-situ formalization'' -- with 13 student collaborators writing the proof assistant code by hand and no AI assistance. Our work differs in that a single mathematician wrote zero code of any kind: the mathematical proof was generated by Gemini, and all \lean\ code was generated by Claude Code from natural-language prompts.

\paragraph{Large-scale autoformalization.}
Math.Inc's Gauss agent \citep{mathinc2026sphere} eliminated $\sim$160K lines of sorry's in the sphere packing formalization (Viazovska's Fields Medal-winning result) in three weeks, growing the codebase from $\sim$22K to 181K lines. The theorem statements and definitions had been written by a team of 6+ human experts; Gauss's role was sorry elimination -- filling in proofs where the goal was unambiguous. \citet{urban2025topology} formalized 130{,}000 lines of Munkres' topology textbook in two weeks for $\sim$\$100, using ChatGPT/Claude Sonnet with a fast proof checker (Megalodon). Both projects are larger in scale but formalize known results; our work targets a new theorem requiring new definitions.

\paragraph{Formalization in physics.}
\citet{toobysmith2026} formalized the stability of the two-Higgs-doublet-model potential in \lean/PhysLean (1.6K lines), discovering an error in a widely cited 2006 paper -- an example of formalization catching bugs in published proofs. Unlike our work, this formalized existing results with minimal AI assistance.

\paragraph{Transparency.}
A distinguishing feature of our work is full transparency. Systems like Seed-Prover \citep{seedprover2025} are closed-source, and even open systems like Numina-Lean-Agent \citep{numina2025} do not publish the complete interaction logs. We publish all 229 human prompts (with timestamps), all 213 commits, the full Gemini dialogue, and the complete development log of all 122 \babysit\ cycles.

\section{Limitations}
\label{sec:limitations}

\paragraph{Mathematical novelty.} The theorem -- that every smooth positive steady state of the full VML system with Coulomb collisions on the torus must be a global Maxwellian -- does not appear in the literature as a stated and proved result. Each proof technique is individually classical: the H-theorem and null-space characterization of the Landau operator are standard \citep{villani2002,cercignani1988}; the polynomial-matching argument that forces constant temperature and Killing's equation on the bulk velocity appears in \citet{cercignani1988} and \citet{desvillettes2005}; and the maximum-principle argument for the nonlinear Poisson--Boltzmann equation is textbook. However, prior work assembles these ingredients only for simpler systems. \citet{desvillettes2005} prove steady-state uniqueness for the spatially inhomogeneous Boltzmann equation \emph{without} electromagnetic fields ($E = B = 0$). \citet{guo2003,guo2012} study the full VML and Vlasov--Poisson--Landau systems but prove \emph{dynamical} stability of small perturbations near an assumed Maxwellian, embedding the relevant macroscopic constraints inside linearized energy estimates rather than stating static uniqueness as a standalone result. What is new is the complete static characterization for the electromagnetically coupled system: the observation that the Lorentz force contributes only $\mathcal{O}(|v|^1)$ terms to the Vlasov equation for a local Maxwellian (because $(v \times B) \cdot v = 0$), so the $\mathcal{O}(|v|^3)$ and $\mathcal{O}(|v|^2)$ constraints of \citet{cercignani1988} and \citet{desvillettes2005} carry over unchanged; the Amp\`ere-law integration argument that forces the bulk velocity to vanish; and the closure via Gauss's law with a neutralizing background. These steps are individually straightforward, but their assembly into a single self-contained proof for the full VML system appears to be new; the complete proof is given in \citet{ilin2026vml}.

\paragraph{Definition alignment.} As discussed in \Cref{sec:lessons}, the definition-alignment bug that survived the entire development process illustrates a broader risk: similar undetected misalignments in other definitions (e.g., \texttt{torusCurlX}, \texttt{LandauOperator}) cannot be ruled out without a thorough expert audit.

\paragraph{Reproducibility.} The process depends on the specific capabilities of Claude Opus 4 (March 2026), Aristotle (March 2026), and Gemini (February 2026). These models are updated frequently; the same process may yield different results with future or past model versions.

\paragraph{Cost.} While the subscription cost was \$200, the API-equivalent cost was $\sim$\$6{,}300 (with prompt caching) or $\sim$\$42{,}700 (without). The gap between the subscription and API cost reflects the Claude Max plan's unlimited usage model. At API pricing without caching, the project would not be as accessible.

\paragraph{Human effort.} Although no code was written, the process required significant mathematical expertise and approximately 50 hours of active supervision (\Cref{tab:stats}). It requires a mathematician who understands the proof well enough to detect when the agent is weakening the theorem. As one Zulip commenter noted, ``the human verification process... is critical'' and cannot yet be replaced by another LLM.

\section{Conclusion}
\label{sec:conclusion}

We have demonstrated that the full loop of AI-assisted mathematical research can be closed in a matter of days: starting from an open conjecture, an AI reasoning model (Gemini DeepThink) correctly resolved the question and produced a detailed proof requiring no corrections; an agentic coding tool (Claude Code) translated this proof into 10K+ lines of \lean~4 from natural-language prompts alone; a cloud-based prover (Aristotle) closed individual lemmas and caught false conjectures; and the \lean\ kernel verified the final result with guaranteed logical correctness. A single mathematician supervised the entire process at a cost of \$200, writing zero lines of code.

The process is not yet fully autonomous -- the human's role in maintaining theorem integrity is essential and cannot currently be delegated to the AI. But the barrier to entry has dropped dramatically: from years of \lean\ expertise to the ability to supervise an AI agent and evaluate theorem statements.

A striking feature of this project is that the formalization was completed \emph{before} the traditional mathematics paper containing the informal proof. The 10-day formalization finished on March 10; the companion math paper, which requires cleaning up the natural-language arguments and obtaining co-author sign-off, is still in preparation. This reversal of the usual order -- where formalization lags publication by months or years -- suggests that AI-assisted formalization has reached a point where it takes comparable effort to produce a machine-checked proof and to write a polished mathematical manuscript.

We believe this workflow -- conjecture, generate proof, formalize, verify, review -- will become a common mode of computer-assisted mathematical research. Each stage of the pipeline is now within reach of current AI systems: reasoning models can resolve conjectures and produce detailed proofs, coding agents can translate these proofs into formal languages, automated provers can close gaps and catch errors, and proof kernels provide the final guarantee of correctness. The mathematician's role shifts from writing proofs to posing questions and evaluating claims -- a supervisor of an AI-driven research pipeline rather than its sole executor.

\bibliography{references}

\end{document}